\documentclass[10pt, a4paper]{article}
\usepackage{hybridreport}
\usepackage[square,authoryear]{natbib}
\usepackage{placeins}
\usepackage{pifont}
\AtBeginDocument{\renewcommand{\checkmark}{\ding{51}}}

\DeclareTextFontCommand{\textbf}{\bfseries}
\let\cite\citep

\newcommand{\ecosys}{AIRoA MoMa}
  \newcommand{\dsid}{airoa-moma-5k}

  \newcommand{\urlDataset}{\url{https://huggingface.co/datasets/airoa-org/airoa-moma-5k}}
  \newcommand{\urlModel}{\url{https://huggingface.co/airoa-org/airoa-pi05-hsr-base}}
\newcommand{\reviewnote}{\footnote{\urlDataset}}
  \newcommand{\rebakenote}{\footnote{\urlRebake; production implementation: \urlRebakeLegacy}}
  \newcommand{\modelnote}{\footnote{Checkpoint: \urlModel; code: \urlCode}}
  \newcommand{\urlCode}{\url{https://github.com/airoa-org/airopi-share}}
  \newcommand{\urlRebake}{\url{https://github.com/airoa-org/rebake}}
  \newcommand{\urlRebakeLegacy}{\url{https://github.com/airoa-org/rebake-legacy}}
  
\newcommand{\sitenote}{}
  \newcommand{\relpages}{the public release pages}

  \newcommand{\teaserfile}{figures/fig1_airoa_river_visual.pdf}
\newcommand{\tabnote}[1]{\par\vspace{3pt}\begin{minipage}{\columnwidth}\footnotesize #1\end{minipage}}

\reporttype{Technical Report}
\logo{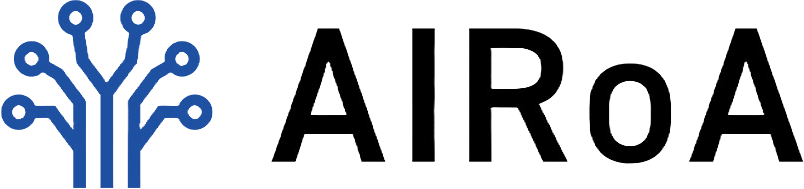}
\title{\ecosys{}: A Fleet-Scale Platform for Mobile-Manipulation Data Collection, Conversion, and VLA Pre-Training}
\shorttitle{\ecosys{}: A Fleet-Scale Platform for Mobile Manipulation}

\author{
  \mbox{Ryosuke Takanami\textsuperscript{1,2*}}\quad
  \mbox{Petr Khrapchenkov\textsuperscript{1*}}\quad
  \mbox{Jumpei Arima\textsuperscript{1,3}}\quad
  \mbox{Takuya Okubo\textsuperscript{2}}\quad
  \mbox{Genki Sano\textsuperscript{4}}\quad
  \mbox{Akinobu Mizutani\textsuperscript{7}}\quad
  \mbox{Tomohiro Motoda\textsuperscript{5}}\quad
  \mbox{Kensuke Chiyoma\textsuperscript{1}}\quad
  \mbox{Masafumi Takahashi\textsuperscript{1}}\quad
  \mbox{Akihito Goto\textsuperscript{1,3}}\quad
  \mbox{Yueh-Hua Wu\textsuperscript{1}}\quad
  \mbox{Hakaru Tamukoh\textsuperscript{7}}\quad
  \mbox{Hiroshi Bito\textsuperscript{3}}\quad
  \mbox{Yukiyasu Domae\textsuperscript{5}}\quad
  \mbox{Yusuke Iwasawa\textsuperscript{1,2}}\quad
  \mbox{Yutaka Matsuo\textsuperscript{1,2}}\quad
  \mbox{Tetsuya Ogata\textsuperscript{1,6}}\quad
  \mbox{Kei Ota\textsuperscript{1}}\quad
  \mbox{Tatsuya Matsushima\textsuperscript{1,2}}
}

\institution{
  \textsuperscript{1}AI Robot Association (AIRoA)\quad
  \textsuperscript{2}The University of Tokyo\quad
  \textsuperscript{3}Toyota Motor Corporation\quad
  \textsuperscript{4}Telexistence, Inc.\quad
  \textsuperscript{5}National Institute of Advanced Industrial Science and Technology (AIST)\quad
  \textsuperscript{6}Waseda University\quad
  \textsuperscript{7}Kyushu Institute of Technology
}

\date{November 2026}
\correspondence{%
  \href{mailto:takanami.ryosuke@airoa.org}{\texttt{takanami.ryosuke@airoa.org}}\\
  Project page: \url{https://research.airoa.io/publications/moma/}\\
  \textsuperscript{*}Equal contribution\\
  This work has been submitted to the IEEE for possible publication. Copyright may be transferred without notice, after which this version may no longer be accessible.%
}

\abstracttext{
Robot policies are often pre-trained on diverse real-world demonstrations
and then fine-tuned for individual tasks. For mobile manipulation tasks
such as shelf restocking, demonstrations must capture both base motion
and object handling. Collecting such data across sites requires consistent
procedures, yet a finished dataset reveals little about how demonstrations
were recorded, checked, and selected for training. We present the
\ecosys{} (Mobile Manipulation) platform, which we used to collect over
80{,}000 hours of teleoperation data with 56 Toyota Human Support Robots
across eight sites. We record the order and outcome of each step so
users can select individual steps while retaining their task context.
We release rebake, which converts ROS recordings into LeRobot episodes,
and \ecosys{} 5k, a dataset of 5{,}025 hours of successful steps.
We continually pre-trained the $\pi_{0.5}$ vision-language-action model
on 15{,}952 hours of internal data and release the checkpoint and training
code. Fine-tuned policies succeeded in 60 of 70 real-robot trials
(85.7\%) across 14 steps from five tasks, each tested from its
designated initial state.\sitenote
}

\begin{document}

\maketitle

\begin{figure}[t]
  \centering
  \includegraphics[width=\textwidth]{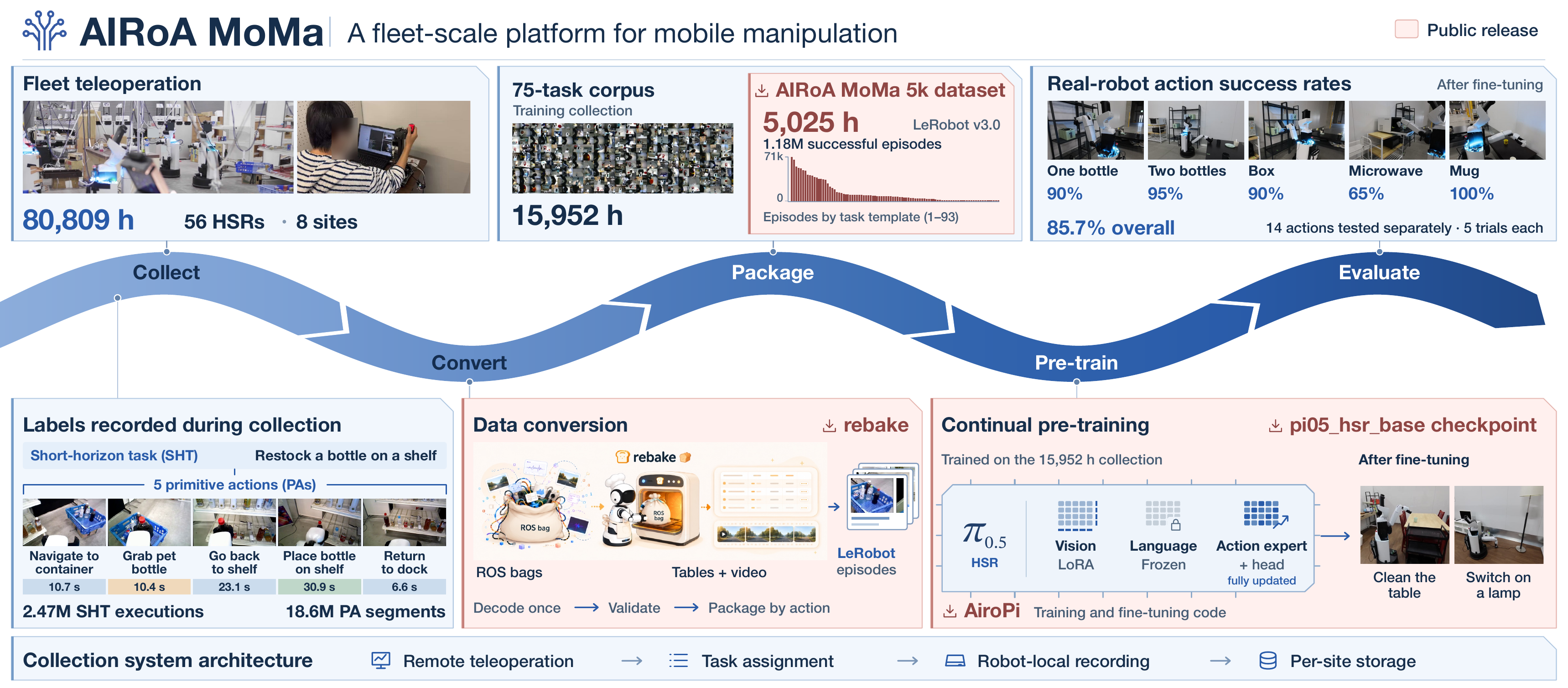}
  \caption{The \ecosys{} platform links fleet collection to real-robot evaluation. Red panels highlight the 5k dataset, conversion software, and policy checkpoint and code. The policy was trained on the separate 15,952-hour collection.}
  \label{fig:teaser}
\end{figure}

\section{Introduction}
\label{sec:introduction}

Vision-Language-Action (VLA) models, which ground natural-language
instructions in real-robot behavior, have emerged as a promising paradigm
for general-purpose robots in everyday human environments~\cite{rt2,openvla,pi0,pi05,geminirobotics}, and their capabilities are bounded
to a large extent by the scale, diversity, and quality of the robot data they
are trained on. Large robot datasets have accordingly broadened the tasks,
environments, and embodiments that policies cover~\cite{oxe,droid,bridgedata,agibotworld}.

Mobile manipulation has shared less in this progress. Some generalist
policies rely on proprietary corpora~\cite{pi0,lingbotvla2,xiaomirobotics1},
and many open collections emphasize tabletop manipulation, whereas
whole-body mobile manipulation requires experience coordinating the base
and arm during the same task~\cite{pi05}. Open datasets also say little
about how they were produced. A collection team must record what happened
during each task, align sensor streams with control commands, and decide
which episodes to retain, and these decisions are difficult to recover from
a finished dataset.

We present the \emph{\ecosys{}} platform (MoMa: \emph{Mo}bile
\emph{Ma}nipulation), which we used to collect fleet data, train policies,
and evaluate them on a real robot (Fig.~\ref{fig:teaser}). We document how the collection
system operated across sites and release the conversion software, a
dataset of mobile-manipulation demonstrations, and a policy checkpoint
with its training code.

During collection, operators perform each task as a predefined sequence
of steps. We call each
such task a \emph{Short-Horizon Task} (SHT), each step a
\emph{Primitive Action} (PA), and the ordered PAs of an SHT its \emph{PA chain}. For example, stocking a shelf includes moving
to the shelf and grasping a bottle. Each recorded PA carries the identifier
of its task execution, its position in the chain, and its outcome
(Sec.~\ref{sec:structure}). Individual PAs can therefore be selected for
training without losing their task context.

Our contributions are as follows:
\begin{itemize}
  \item \emph{Fleet design and operating procedures.} We describe how we
  deployed THSR~\cite{wakayama2025thsr}, a joint-mapped leader--follower
  teleoperation system, on 56 Toyota Human Support Robots (HSRs) across eight sites and how we
  managed data quality while recording over 80{,}000 hours (Sec.~\ref{sec:collection}).
  \item \emph{Conversion software.} We release rebake, the software used
  to convert the fleet's ROS recordings into episodes in the LeRobot dataset format~\cite{lerobot}. Its
  intermediate records can be queried, and robot-specific settings are
  confined to one configuration file (Sec.~\ref{sec:rebake}).
  \item \emph{Demonstration data.} \ecosys{} 5k contains 1{,}184{,}259
  successful PA episodes (5{,}025 hours) from five sites in LeRobot v3.0
  format. Each episode identifies its task, position, site, and robot.
  The release has 53.5 times the recorded hours of our
  preliminary single-site dataset~\cite{airoamoma} (Sec.~\ref{sec:datasets}).
  \item \emph{Policy checkpoint and training code.} We release the
  $\pi_{0.5}$ checkpoint~\cite{pi05} obtained by continual pre-training (CPT)
  on 15{,}952 hours from the fleet, using low-rank adaptation
  (LoRA)~\cite{lora} for the vision encoders (Sec.~\ref{sec:recipe}).
  Policies fine-tuned from it achieved 85.7\% PA-level success across five
  real-robot tasks and performed longer household tasks
  (Sec.~\ref{sec:casestudy}).
\end{itemize}
 \section{Related Work}
\label{sec:related_work}

\subsection{Real-Robot Data for Generalist Policies}
Generalist robot policies learn from demonstrations, so the progression
from RT-2~\cite{rt2} through open models such as OpenVLA, RDT-1B, and
GR00T~N1~\cite{openvla,rdt1b,groot_n1} to flow-matching VLAs such as
$\pi_0$ and $\pi_{0.5}$~\cite{pi0,pi05} and proprietary systems such as
Gemini Robotics~\cite{geminirobotics} has been driven as much by data as by
architecture. The open datasets behind this progress scale along
complementary axes. Open X-Embodiment (OXE)~\cite{oxe} aggregates over one
million trajectories from 22 embodiments, DROID~\cite{droid} spans 76k
trajectories and 564 scenes, and BridgeData V2~\cite{bridgedata} emphasizes
task and environment diversity. These datasets broaden the range of tasks and robots available for
learning. \ecosys{} 5k complements them with demonstrations in which a mobile
base and an arm perform successive steps of the same task.

\subsection{Mobile-Manipulation Data and How It Is Collected}
Mobile manipulation is collected by teleoperation, because contact-rich
behaviors such as opening doors or loading appliances are hard to script or
plan~\cite{mobilealoha}. Whole-body teleoperation devices have made
this practical. Mobile ALOHA, THSR, and the BEHAVIOR Robot Suite each pair
a mobile manipulator with a leader device~\cite{mobilealoha,wakayama2025thsr,behaviorrobotsuite}. Handheld
grippers such as UMI~\cite{umi} and YUBI~\cite{yubi} remove the robot from
collection altogether but record only the hand, so whole-body behavior
still has to be collected on the robot. Teleoperated datasets are
growing. BRMData~\cite{brmdata} and Galaxea~\cite{galaxea} cover
bimanual-mobile household manipulation, and AgiBot
World~\cite{agibotworld} and RoboMIND 2.0~\cite{robomind} scale whole-body
and multi-embodiment collection to $10^5$--$10^6$ trajectories. Other
recent releases add force and audio (RH20T~\cite{rh20t}) or further scale
(ABC-130K and HIW-500~\cite{abc130k,hiw500}).
Table~\ref{tab:datasets} compares them with \ecosys{} 5k on released
hours, a mobile embodiment, synchronized force sensing, sub-task structure,
and the release of conversion tooling and a trained model. \ecosys{} 5k is the only entry that
provides all of these, and it does so on a single embodiment and action
schema, so that all of its hours can be pooled directly.

\subsection{From Recordings to Training Data}
Every one of these efforts had to turn raw robot recordings into a
training-ready format, yet this step is rarely released
(Table~\ref{tab:datasets}, Tools).
LeRobot~\cite{lerobot} standardizes the target format, and AgiBot
World~\cite{agibotworld} describes its collection system, but the
conversion itself (synchronizing multi-rate streams, checking signal
validity, segmenting episodes, and gating quality) is typically left to ad
hoc, robot-specific scripts that each group rewrites. We release this step as
rebake (Sec.~\ref{sec:rebake}) so that other ROS-recorded robots can reuse
it.

\begin{table}[t]
  \centering
  \caption{Real-robot datasets by released hours and features.}
  \label{tab:datasets}
  \small
  \setlength{\tabcolsep}{3.6pt}
  \begin{tabular}{@{}lrccccc@{}}
    \toprule
    Dataset & Rel.\ h & Mobile & F/T & Hier. & Tools & Model \\
    \midrule
    OXE/RT-X~\cite{oxe}              & ---      & mixed & --- & --- & \checkmark & \checkmark \\
    DROID~\cite{droid}               & 350      & no    & --- & --- & --- & \checkmark \\
    BridgeData V2~\cite{bridgedata}  & ---      & no    & --- & --- & --- & --- \\
    RH20T~\cite{rh20t}               & ---      & no    & \checkmark & --- & --- & --- \\
    RoboMIND 2.0~\cite{robomind}     & $>$1{,}000 & mixed & --- & --- & --- & --- \\
    AgiBot World~\cite{agibotworld}  & 2{,}976  & yes   & --- & \checkmark & \checkmark & \checkmark \\
    Galaxea~\cite{galaxea}           & 500      & yes   & --- & \checkmark & --- & \checkmark \\
    ABC-130K~\cite{abc130k}          & $\sim$3{,}500 & no & --- & part. & --- & --- \\
    HIW-500~\cite{hiw500}            & $>$500   & yes   & --- & --- & --- & --- \\
    \textbf{\ecosys{} 5k (ours)}    & \textbf{5{,}025} & \textbf{yes} & \checkmark & \checkmark & \checkmark & \checkmark \\
    \bottomrule
  \end{tabular}
  \tabnote{Values for other datasets are taken from their publications
  (``---'': not reported or not released). Rel.\ h: released hours. Mobile:
  yes = collected on mobile manipulators; mixed = multi-embodiment aggregate
  that includes fixed-base robots. F/T: synchronized force--torque. Hier.:
  sub-task-level annotations (part.: subset of episodes). Tools: open-source
  conversion tooling released. Model: a released policy trained on the
  dataset or on the collection it is drawn from.}
\end{table}

\subsection{Task Structure and Contact Signals}
Beyond scale, two properties of the data determine what can be learned from
it. The first is task structure. Long-horizon activities contain reusable
behaviors whose outcomes determine later states.
BEHAVIOR-1K~\cite{behavior1k} formalizes this structure in simulation, and
$\pi_{0.5}$~\cite{pi05} exploits it by predicting semantic subtasks before
low-level actions. Existing real-robot datasets expose this structure, if
at all, as sub-task annotations on a subset of episodes
(Table~\ref{tab:datasets}, Hier.). The second is contact. Force signals
reveal states that are weakly observable in images, and prior work learns
visual--tactile representations~\cite{visiontouch}, grounds heterogeneous
sensors through language~\cite{fuse}, or adds force-aware experts to
VLAs~\cite{forcevla}. The \ecosys{} platform records both during collection. Every PA
segment carries its position in the task and its success label, in
chains of up to 21 PAs (Sec.~\ref{sec:structure}), together with
synchronized force--torque histories and per-servo telemetry
(Sec.~\ref{sec:schema}). A user can therefore train on single PAs, model
whole SHTs, hold out combinations of PAs, or study contact.
 \section{Collection System}
\label{sec:collection}

Collecting mobile-manipulation demonstrations across sites requires a
consistent way to operate the robots and record their actions. We describe
the robot and its teleoperation device (Sec.~\ref{sec:platform}), the
fleet architecture (Sec.~\ref{sec:architecture}), the task structure
recorded during collection (Sec.~\ref{sec:structure}), and the defects we
removed at the source (Sec.~\ref{sec:quality}).
Together the recordings form the \emph{raw pool} before curation
(Table~\ref{tab:tiers}). The conversion software in Sec.~\ref{sec:rebake}
processes this pool into the datasets described in Sec.~\ref{sec:datasets}.

\subsection{Robot and Teleoperation}
\label{sec:platform}
\begin{figure}[t]
  \centering
\includegraphics[width=0.6\textwidth]{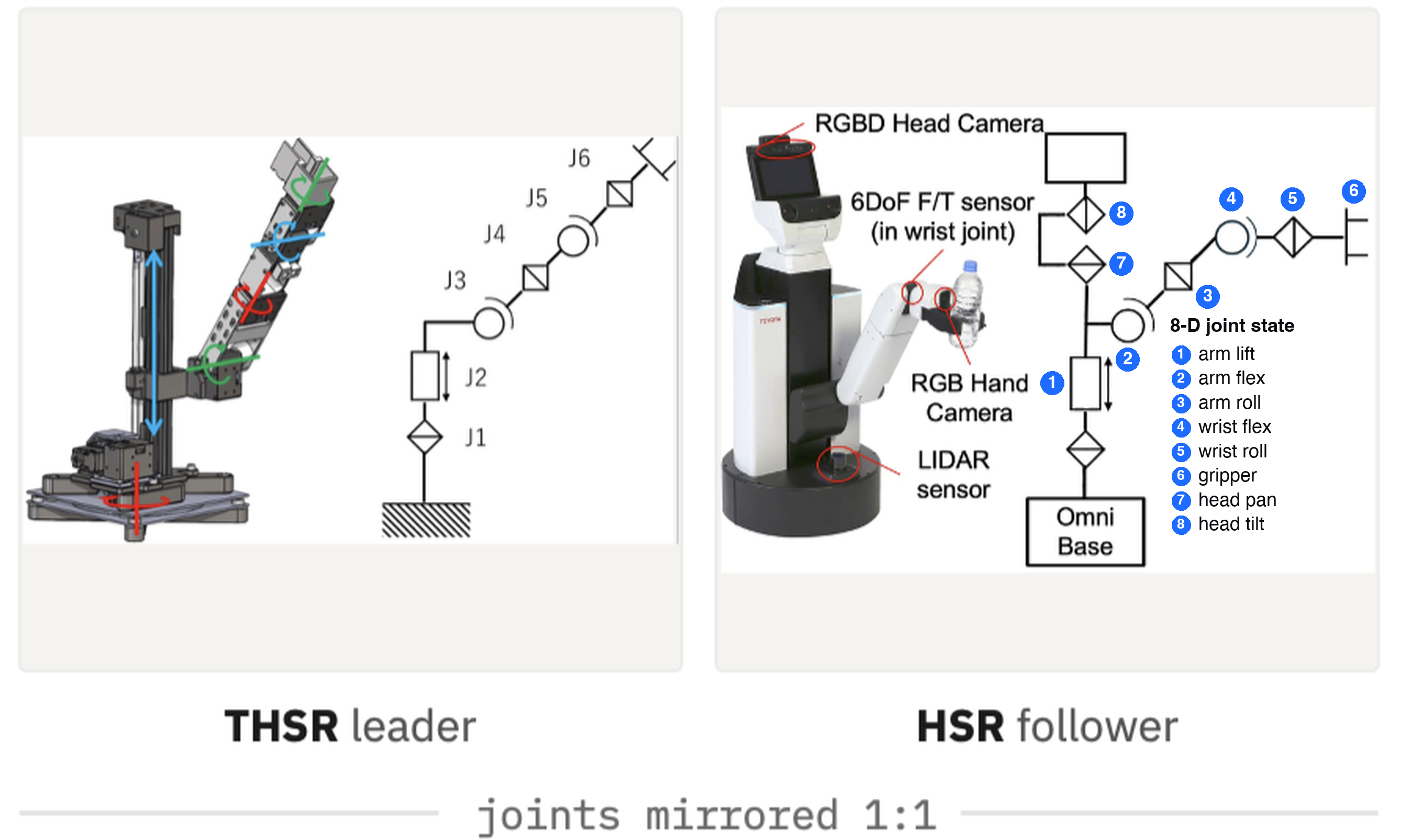}
  \caption{The THSR leader (left) and the HSR follower (right). The
  leader's joints map one-to-one onto HSR's arm joints, lift axis, and base
  rotation; HSR carries an RGBD head camera, an RGB hand camera, and a
  six-axis force--torque sensor in the wrist on an omnidirectional base.
  Numbers mark the eight joints of the released joint state
  (Sec.~\ref{sec:schema}).}
  \label{fig:thsr}
\end{figure}
The platform collects data with the Toyota HSR~\cite{hsr} in household, retail, kitchen, and warehouse settings. The tasks focus on
behaviors that tabletop corpora underrepresent, such as navigation interleaved with
manipulation, articulated mechanisms (doors and trays),
tight-clearance placement, repeated multi-object handling, and restoring
items that have fallen out of place (Fig.~\ref{fig:teaser}). Object
placement, lighting, and the robot's initial pose vary across episodes.
HSR is a mobile service robot~\cite{hsr} with a four-degree-of-freedom (DoF) arm on a lift axis, a 1-DoF gripper, a 2-DoF head, and a 3-DoF omnidirectional base,
all commanded together. Because every unit is the same robot model, all
episodes share one state and action format. The 56 units and eight sites
still vary in hardware, operators, and environment.

Demonstrations are collected with THSR~\cite{wakayama2025thsr}, a
leader--follower device that maps the operator's joints one-to-one onto
HSR (Fig.~\ref{fig:thsr}). We chose it because the alternatives do
not suit contact-rich manipulation on HSR. The default gamepad interface
commands one axis at a time, and end-effector-space devices such as VR
controllers require inverse kinematics (IK) over arm, lift, and base,
since the arm alone has fewer than six degrees of freedom. THSR instead
transmits the leader's joint states directly as joint-position commands to
HSR's arm joints, lift axis, and base-rotation axis, so the follower
mirrors the operator's posture without solving IK. Base translation, head
motion, and the gripper are controlled separately with a Joy-Con
controller. The mechanical design of the leader is described
in~\cite{wakayama2025thsr}.

Operators complete proficiency training before they collect data. Each of
their sessions is recorded as ROS bags, the native logging format of ROS,
and converted offline (Sec.~\ref{sec:rebake}).

\begin{figure}[t]
  \centering
\includegraphics[width=\textwidth]{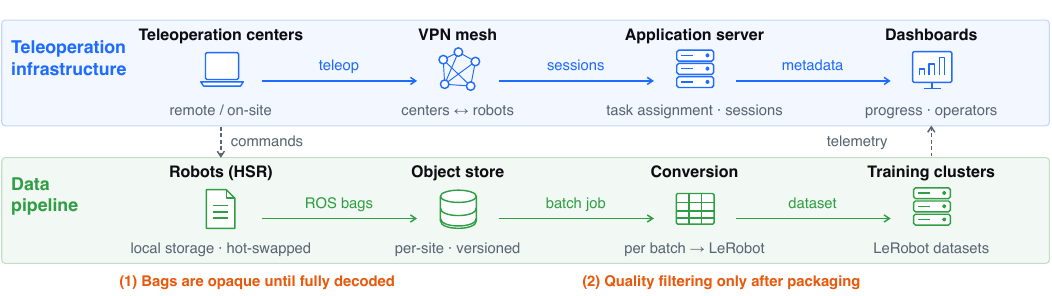}
  \caption{Fleet architecture of the collection system: the teleoperation
  infrastructure (teleoperation centers over a mesh VPN, task-assignment
  server, dashboards) and the data pipeline (robot-local storage, per-site
  object store, batch conversion, training clusters). Markers (1) and (2) are the two bottlenecks that the
  robot-agnostic conversion of Sec.~\ref{sec:rebakerust} removes: bags
  opaque until fully decoded, and quality filtering only after packaging.}
  \label{fig:arch}
\end{figure}

\begin{figure}[t]
  \centering
\includegraphics[width=\textwidth]{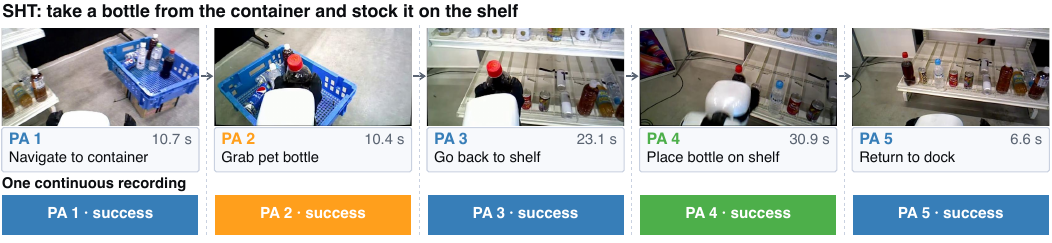}
  \caption{An example SHT execution (``take a bottle from the container and
  stock it on the shelf''): one head-camera frame per PA (top) and the same PAs as
  contiguous, success-labeled windows of one continuous recording (bottom),
  ordered by their position in the chain. Colors mark PA families
  (navigation, grasp, placement) and are shared with
  Fig.~\ref{fig:taskcards}.}
  \label{fig:taskstructure}
\end{figure}

\subsection{Scalable Fleet Architecture}
\label{sec:architecture}
Operating many robots at several sites requires infrastructure beyond the
teleoperation device. The architecture has four components
(Fig.~\ref{fig:arch}). (i)~\emph{Remote
teleoperation centers.} Operators work from centers connected to the robots over a mesh VPN, so
that operator recruitment is decoupled from robot location.
(ii)~\emph{Collection application and backend.} The application shows
operators live sensor feeds while a centralized backend coordinates task
assignment, session tracking, and upload orchestration. (iii)~\emph{Robot-local, compressed recording.} Recordings are written to
storage physically attached to the robot, because streaming over wireless
links drops frames under network jitter, and camera streams are
JPEG-compressed, which cuts per-episode storage by an order of magnitude
with no visible loss in image quality. When a local drive fills, it is
hot-swapped so that the session continues uninterrupted. (iv)~\emph{Upload and provenance.} An agent uploads recordings with
automatic retry to a versioned, site-partitioned object store, and
dashboards track progress and operator performance.

\subsection{SHT/PA Structure Recorded During Collection}
\label{sec:structure}
Operators execute templated SHTs. An \emph{SHT template} names a task
(``take a bottle from the container and stock it on the shelf'') and fixes
its canonical chain of PAs. PA boundaries and outcomes are recorded
during the session, so the pool is structured during recording rather than
annotated post hoc (Fig.~\ref{fig:taskstructure}). The filters of Sec.~\ref{sec:rebakelegacy} can therefore act on each PA
segment. Each recorded PA segment carries a natural-language PA label, the parent
SHT template and its position in that template's chain, an execution
identifier (\texttt{uuid}) shared by the PAs of one SHT execution, and the
collection site, robot identifier, and success label. These fields
make the data usable without re-annotation. The identifier and position let
the PAs of one execution be reassembled into the full task, the success
label lets curation accept or reject each PA on its own, and the site and
robot identifiers let a user hold out a site or a robot.

The pool holds more than 80{,}000 recorded hours from eight sites,
organized into 2.47 million SHT executions and 18.6 million labeled PA
segments that cover 99.7\% of the recorded time (Table~\ref{tab:tiers}).
The five sites in the released datasets contribute 97\% of it. In the pool, tasks come
from home and retail settings in roughly equal volume, and the time is
concentrated in a few labels. Of the 1{,}432 task labels (the task names recorded with
the sessions), 205 cover 92\% of the time, which makes the task-level
curation of Sec.~\ref{sec:training75} possible.

\subsection{Defects Removed at the Source}
\label{sec:quality}
The filters of Sec.~\ref{sec:rebakelegacy} discard bad episodes after
recording. Some defects are better removed at the source, because a defect
fixed there disappears from every dataset converted afterwards. We monitor
signal integrity (dropouts, frame loss), kinematic quality (hesitation,
micro-tremor), and semantic consistency (instruction--trajectory mismatch),
and removed four such defects. Robot-local recording
(Sec.~\ref{sec:architecture}) removed frame drops under network jitter,
and clipping and binarization removed noise in the gripper channels.
Automated cross-modal checks catch instruction--trajectory mismatch, and a
per-robot correction removes encoder offsets that made policies drift
slowly.
 \section{Data Conversion: rebake}
\label{sec:rebake}

Every dataset and model in this paper depends on converting the recorded
bags into training-ready datasets. We release this conversion in full as \emph{rebake}, an Apache-2.0
bag-to-LeRobot pipeline linked from \relpages{}. The implementation that
produced the datasets of this paper is archived alongside it.\rebakenote{}

\subsection{Pipeline}
\label{sec:rebakelegacy}
Each uploaded batch of bags
is converted into an independent, version-tagged LeRobot dataset, so that a
training package can be assembled from any subset of batches. The pipeline
pairs each bag with the metadata recorded during the session (task label,
PA boundaries, success labels, site, and robot) and verifies these
annotations rather than creating them post hoc. Quality is enforced in
stages, because each stage catches a different defect at the lowest cost.
A session is first checked for signal validity (no dropped or stalled
streams) and for motion (the robot actually moved), and is then cut at the recorded boundaries into PA
episodes, one training episode per PA segment. A three-stage filter then gates the
output at the SHT-execution level (a success label is present and the
duration is not an outlier for that task), the PA level (the same criteria
per segment), and the signal level (joint-position statistics, repeated
camera frames, and per-episode outliers in the signal channels), and random
samples are inspected visually as a final check. Streams are synchronized
at 10~Hz, a rate that keeps the corpus tractable while the 100~Hz force
and joint histories are preserved as per-frame windows
(Sec.~\ref{sec:schema}). Images are stored as compressed video and numeric
channels as columnar tables. Because synchronization carries a value
forward when a stream has not updated, every channel also stores a
\emph{freshness indicator} that tells a user whether the value at a given
step was newly received or repeated.

\subsection{Robot-Agnostic Conversion}
\label{sec:rebakerust}
Converting bags directly into a dataset has two bottlenecks that limit how
quickly new robots or new dataset variants can be produced
(Fig.~\ref{fig:arch}). (1)~A ROS~1 bag is opaque until fully
decoded, so nothing can be validated before the whole message stream has
been read. (2)~Quality filtering can run only after packaging, so
problems surface after expensive processing. rebake
removes both by parsing each ROS~1 bag, ROS~2 bag, or MCAP file (the ROS~2
logging container) once into a columnar intermediate of tables and encoded
video that can be queried like a database. Validation then runs on the
intermediate before anything is packaged, and LeRobot datasets can be generated
repeatedly from it with different task filters, rates, or feature subsets
without touching the original bags. Multi-rate synchronization,
coordinate-frame transforms, and video encoding are built-in stages, and for
robots whose topics use supported message types, adding a robot requires
only a declarative configuration file that maps its ROS topics to LeRobot
features, which makes the converter reusable for other ROS-recorded robots.

\subsection{Privacy Processing for Release}
\label{sec:privacy}
The head and hand cameras record bystanders, whose faces must not be
released. We therefore run a YOLO-based person detector on every video and
release in blurred form those in which the detection confidence exceeds a
preset threshold (8.6\% of videos). No episodes are dropped.
 \section{Datasets}
\label{sec:datasets}

We produced two datasets from the raw pool of Sec.~\ref{sec:collection}
with the tooling of Sec.~\ref{sec:rebake}. \emph{\ecosys{} 5k} is the
public release, and the \emph{training collection} is the internal set on
which the released checkpoint was trained. Sec.~\ref{sec:training75} describes how both are derived and what the
training collection contains (Table~\ref{tab:tiers}), and
Secs.~\ref{sec:schema} and~\ref{sec:release} describe the public release.

\subsection{From the Raw Pool to the Release}
\label{sec:training75}
We consolidate the 1{,}432 task labels in the pool into 75 tasks, and the
numbered initial-condition variants of a task (for example, different
starting shelves) give 93 SHT templates. These 75 tasks account for about
38{,}000 of the 80{,}809 recorded hours. During curation, we screened each
PA episode for timestamp delays, abrupt changes in joint position,
durations outside the expected range for the task, and corrupted video.
We excluded episodes that failed these quality checks. The internal
collection used for continual pre-training (Sec.~\ref{sec:experiments})
comprises 4{,}046{,}550 PA episodes totaling 15{,}952 hours. Of these,
2{,}484 episodes (0.06\%) are labeled as failures (Table~\ref{tab:tiers}).
\ecosys{} 5k, the public artifact (Sec.~\ref{sec:release}), is the
\emph{core} drawn from the same 75 tasks. It spans the same 93 SHT
templates, includes recordings made after the training collection was
frozen, and retains only the successful episodes that pass stricter
thresholds on duration and per-episode signal statistics, so that a user
downloads the cleanest subset first.

\begin{table}[t]
  \centering
  \caption{The raw pool and the two datasets curated from it.}
  \label{tab:tiers}
  \small
  \setlength{\tabcolsep}{4.5pt}
  \begin{tabular}{@{}lrrr@{}}
    \toprule
    & Raw pool & Training collection & \ecosys{} 5k \\
    & internal & internal & public \\
    \midrule
    SHT executions & 2{,}474{,}213 & 692{,}785 & 425{,}518 \\
    PA segments/episodes & 18{,}580{,}436 & 4{,}046{,}550 & 1{,}184{,}259 \\
    Frames & --- & 574.3M & 180.9M \\
    Hours & 80{,}809 & 15{,}952 & 5{,}025 \\
    Task labels / curated tasks & 1{,}432 & 75 & 75 \\
    SHT templates & --- & 93 & 93 \\
    PA labels & 3{,}085 raw & 531 defined & 358 observed \\
    Failed share & 0.8\% & 0.06\% & 0\% (success-only) \\
    Sites / robots & 8 / 56 & 5 / 44 & 5 / 44 \\
\bottomrule
  \end{tabular}
  \tabnote{Raw values count SHT executions and PA segments as recorded, with
  hours recorded to bags. Collection and core
  values count retained PA episodes, with hours summed over episodes; their
  SHT executions are the distinct execution identifiers (\texttt{uuid})
  among retained episodes. Failed share is per session (raw) and per PA
  episode (collection and core). PA labels: raw instruction strings, the
  labels defined by the templates, and the labels observed in the core.}
\end{table}

\begin{figure}[t]
  \centering
\includegraphics[width=\textwidth]{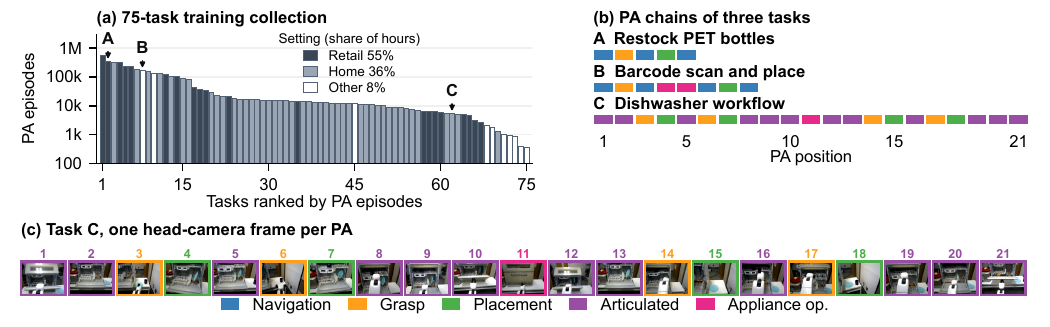}
  \caption{The 75-task training collection at a glance. (a)~PA-episode
  volume is long-tailed across the 75 tasks; shading marks the setting, and
  A--C mark the tasks of~(b). (b)~The PA chains of three tasks, as fixed by
  their templates; colors mark PA families as in
  Fig.~\ref{fig:taskstructure}. (c)~One head-camera frame per PA of the
  dishwasher workflow, the longest chain (21 PAs).}
  \label{fig:taskcards}
\end{figure}

The training collection partitions this experience into 75 task datasets
in retail, warehouse, appliance, household, kitchen, and tabletop settings.
Retail tasks account for 55\% of its hours and home tasks (household,
kitchen, and appliance) for 36\% (Fig.~\ref{fig:taskcards}a).
Grouping its PA episodes by execution identifier yields 692{,}785 SHT
executions, 62.4\% of which contain the full PA chain of their template. Long-horizon structure is exposed as individually addressable segments
rather than long videos, so that a user can train on single PAs or
reassemble chains without re-segmenting video. Of the 75 template chains, 35 contain at least six PAs, 22 contain
at least ten, and the longest, a dishwasher workflow, contains 21 (Table~\ref{tab:tasks75}; Fig.~\ref{fig:taskcards}c).

\begin{table}[t]
  \centering
  \caption{Representative tasks of the 75-task training collection.}
  \label{tab:tasks75}
  \small
  \setlength{\tabcolsep}{5.8pt}
  \begin{tabular}{@{}lrrr@{}}
    \toprule
    Task (category) & PA ep. & Hours & Chain \\
    \midrule
    Items fallen inside shelf (recovery)   & 561{,}437 & 2{,}222 & 6 \\
    Restock 500--600\,ml bottles (restock) & 357{,}387 & 1{,}602 & 5 \\
    Tidy up objects on the table (house.)  & 323{,}100 & 842     & 15 \\
    Return bottles to container (pickup)   & 321{,}566 & 1{,}288 & 5 \\
    Barcode scan, shelf to table (wareh.)  & 173{,}242 & 685     & 8 \\
    Heat an item in the microwave (appl.)  & 155{,}109 & 876     & 10 \\
    Make coffee (appliance)                & 12{,}808  & 70      & 15 \\
    Dishwasher workflow (appliance)        & 5{,}561   & 45      & 21 \\
    Relocate coffee bottles (tabletop)     & 2{,}187   & 17      & 4 \\
    \midrule
    All 75 tasks & 4.05M & 15{,}952 & 2--21 \\
    \bottomrule
  \end{tabular}
  \tabnote{Hours sum episode durations. Chain is the length of the
  template's PA chain.}
\end{table}

High-volume logistics tasks provide dense repetition over many
robots and initial states, while lower-volume appliance tasks contribute the
longest articulated chains (Fig.~\ref{fig:taskcards},
Table~\ref{tab:tasks75}).

\subsection{Observation and Action Schema}
\label{sec:schema}
All released channels are aligned at 10~Hz (native rates: RGB 30~Hz, joint
states and wrist wrench 100~Hz). Each frame provides head and hand
RGB images ($640\times480$), an 8-D joint state, an 11-D relative action, end-effector poses, and raw
teleoperation signals. The 8-D state concatenates the arm lift, arm flex, arm roll, wrist flex,
wrist roll, gripper, head pan, and head tilt joints (numbered in Fig.~\ref{fig:thsr}). The 11-D relative
action expresses the arm and head targets as deltas with respect to the
current state, keeps the gripper command absolute, and appends planar base
increments $(\Delta x,\Delta y,\Delta\theta)$. An absolute base state or action is not provided owing to limited
wheel-odometry accuracy; base motion appears only as relative increments
derived from the teleoperation commands.

The force channel is a 100-sample history of the six-axis wrist wrench
$(F_x,F_y,F_z,M_x,M_y,M_z)$, flattened to 600 dimensions and expressed in the wrist-mounted sensor frame (in 4{,}608 episodes only one
sample per frame was recorded and is repeated to fill the window).
For each of the 11 servo axes (the eight joints above and the three base
drive joints), the release also provides commanded and measured position and velocity,
effort (torque), current, and temperature, which expose contact and load without an
external sensor. Absolute, state-difference, and relative action channels
are provided in parallel so that users of different action
parameterizations need not re-derive them.

\subsection{The Public Release: \ecosys{} 5k}
\label{sec:release}
We release the core as \emph{\ecosys{} 5k}. It contains 1{,}184{,}259 successful PA
episodes (180.9 million frames, 5{,}025 hours) from 44 HSR units at five
sites (LeRobot v3.0, public identifier \texttt{\dsid}) and is available from
\relpages{}.\reviewnote{} To our
knowledge, this is the largest publicly accessible real-robot
mobile-manipulation research dataset by summed released-episode hours
(Table~\ref{tab:datasets}). The LeRobot v3.0 layout~\cite{lerobot} lets
metadata be inspected before downloading the 4-TiB corpus. Because curation accepts or rejects each PA episode on its own, an SHT
execution in the core often lacks some of its PAs. Grouping episodes by
execution identifier gives 425{,}518 executions, of which 2.6\% contain
the full chain of their template (62.4\% in the less strictly filtered
training collection). Users who need complete chains can select them with
the identifier. The core spans all 93 SHT templates and 358 of the 531 PA
labels those templates define (Table~\ref{tab:tiers}).

\begin{figure}[t]
  \centering
\includegraphics[width=0.6\textwidth]{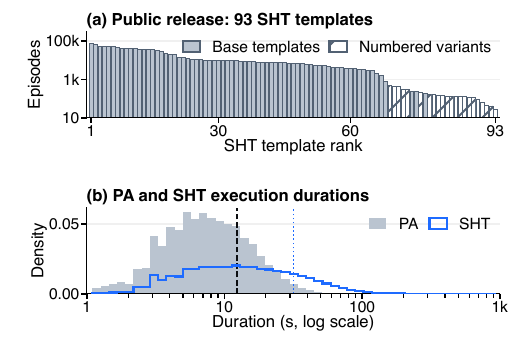}
  \caption{Statistics of \ecosys{} 5k. (a)~Episodes are long-tailed
  across the 93 SHT templates (largest: 71{,}258); hatched bars are the
  initial-condition variants of Sec.~\ref{sec:training75}. (b)~Durations of
  PA episodes and of SHT executions, the latter summed over the PA episodes
  of an execution that are in the release; dashed lines mark the medians,
  12.4~s and 31.7~s.}
  \label{fig:corestats}
\end{figure}

Task volume is strongly long-tailed (Fig.~\ref{fig:corestats}a). The most frequently recorded SHT template has 71{,}258 PA episodes, so training on the release calls for task resampling, as in the recipe of Sec.~\ref{sec:recipe}. A PA episode lasts a median of 12.4~s (Fig.~\ref{fig:corestats}b).

Volume is also long-tailed across the five collection sites. The two
largest contribute 60.7\% and 30.6\% of PA episodes and the remaining
three under 9\% together. Because every episode retains its site and
robot, the release supports held-out-site evaluation.

The base moves in most of the release. By label, 35.7\% of PA episodes are
navigation PAs, and 48 of the 93 templates interleave
navigation and manipulation PAs. The base also moves within manipulation
PAs. In 67.7\% of them the measured drive-wheel velocity exceeds
0.25~rad/s in at least one frame, mostly as brief repositioning
(median 2.2~s) while the arm is moving.
 \begin{figure}[t]
  \centering
\includegraphics[width=\textwidth]{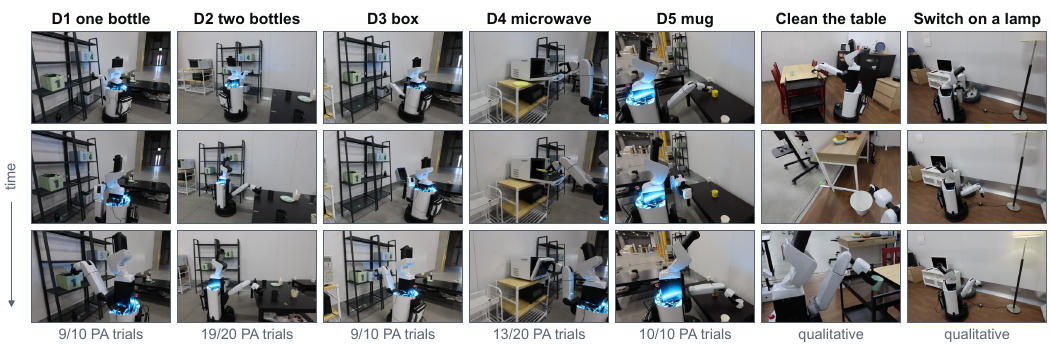}
  \caption{Real-robot evaluation of policies fine-tuned from the released
  checkpoint. Each column shows one task as three frames in time order from
  top to bottom. D1--D5 are the downstream tasks (picking one bottle, picking
  two bottles, picking a box, operating a microwave, and handling a mug),
  with the successful PA trials out of all PA trials below each column (five
  per PA; Table~\ref{tab:eval}). The two right columns are long-horizon
  household tasks, cleaning a table and switching on a floor lamp, shown as
  qualitative examples.}
  \label{fig:eval}
\end{figure}

\section{Continual Pre-Training of \texorpdfstring{$\pi_{0.5}$}{pi0.5}}
\label{sec:experiments}

We release a reusable starting point for HSR policies. It is a
$\pi_{0.5}$~\cite{pi05} checkpoint continually pre-trained on the internal training collection of
Sec.~\ref{sec:training75}, released with its training and fine-tuning
code on \relpages{}.\modelnote{}
We validate it on whole-body mobile manipulation, in which the policy
commands the base, lift, arm, head, and gripper jointly.

\subsection{Training Recipe}
\label{sec:recipe}
Starting from the public $\pi_{0.5}$ base checkpoint~\cite{pi05}
(PaliGemma backbone with a Gemma-2B language model, 300M-parameter action
expert), we perform continual pre-training (CPT) on the training collection with the model's
flow-matching objective for 250k updates at a global batch size of 2{,}048
on 64 NVIDIA H200 GPUs. The language backbone is frozen to preserve its instruction grounding. The
action expert and action head are fully updated because the embodiment is
new. The per-camera (head and hand) vision encoders receive LoRA
adapters (rank 16, $\alpha=16$, on patch embedding, attention, MLP, and
head), which adapt the visual features to the HSR cameras with few
trainable parameters.
Actions use the 11-D relative representation of Sec.~\ref{sec:schema} with a horizon of 16.
Optimization uses cosine decay from a peak learning rate of
$2.5\times10^{-5}$ with 1{,}000 warm-up steps and an exponential moving average (EMA) of the
weights with decay 0.99. The 75 tasks are sampled with equal probability so
that the long tail of Fig.~\ref{fig:taskcards} does not dominate training. The model consumes RGB and proprioception only.

\subsection{Validation: Whole-Body Mobile Manipulation on a Real Robot}
\label{sec:casestudy}
We fine-tune the checkpoint on five real-robot downstream tasks. As a
reference, we also fine-tune the public $\pi_{0.5}$ checkpoint directly on
the same downstream demonstrations. The five tasks (D1--D5 in
Table~\ref{tab:eval}) are picking one or two bottles, picking a box,
operating a microwave, and handling a mug, each decomposed into two to four
PAs. Each PA is evaluated independently
from a designated initial state, so the experiment measures PA-level
execution rather than end-to-end SHT completion (Fig.~\ref{fig:eval}).

\begin{table}[t]
  \centering
  \caption{PA-level successes over trials on the five downstream tasks.}
  \label{tab:eval}
  \small
  \begin{tabular}{@{}lcrr@{}}
    \toprule
    Task & PAs & Direct FT & CPT$\to$FT \\
    & & (200k updates, 3 trials/PA) & (50k updates, 5 trials/PA) \\
    \midrule
    D1 one bottle   & 2 & 6/6   & 9/10 \\
    D2 two bottles  & 4 & 12/12 & 19/20 \\
    D3 box          & 2 & 4/6   & 9/10 \\
    D4 microwave    & 4 & 10/12 & 13/20 \\
    D5 mug          & 2 & 3/6   & 10/10 \\
    \midrule
    All & 14 & 35/42 (83.3\%) & 60/70 (85.7\%) \\
    \bottomrule
  \end{tabular}
  \tabnote{Direct FT: fine-tuning directly from the public $\pi_{0.5}$
  checkpoint. CPT$\to$FT: fine-tuning from the released CPT checkpoint.}
\end{table}

Table~\ref{tab:eval} lists the PA-level outcomes and Fig.~\ref{fig:eval}
the evaluation scenes. Direct fine-tuning produced 35 successes in 42 trials
(83.3\%), and fine-tuning from the continually pre-trained checkpoint
produced 60 successes in 70 trials (85.7\%).
The largest
per-task difference is on D4 (microwave), where the second and third of
its four PAs succeeded in 1 and 2 of 5 trials in the CPT$\to$FT run. The two runs differ in fine-tuning budget (200k vs.\ 50k
updates) and in trials per PA (3 vs.\ 5), so we report them as two
validations of the recipe rather than as a controlled comparison.

As qualitative examples, policies fine-tuned from the checkpoint in the
same way also execute long-horizon household tasks that chain navigation
and manipulation (cleaning a table, switching on a floor lamp;
Fig.~\ref{fig:eval}, right). Together with the 60 of 70 successful PA
trials, these rollouts indicate that a VLA trained through the platform is
usable for whole-body mobile manipulation on HSR.

 \section{Conclusion}
\label{sec:conclusion}

We used the \ecosys{} platform to collect more than 80{,}000 hours of mobile-manipulation
data, convert the recordings into training episodes, and evaluate adapted
policies on a real robot. We document the fleet architecture and operating
procedures and release rebake for converting ROS recordings, \ecosys{} 5k
for training on demonstrations, and a $\pi_{0.5}$ checkpoint with training
code for adaptation to new tasks. The checkpoint was continually pre-trained
on a 15{,}952-hour collection from the fleet.

\textbf{Limitations.} The released core is success-only; failed executions
remain in the raw pool. The training collection behind the checkpoint is internal, so the continual pre-training run
is reproducible at the recipe level, not the data level. The corpus is single-embodiment and long-tailed, with templated
instructions.

\textbf{Future directions.} Held-out-site, compositional, and contact-aware
studies can exploit the release's task structure and provenance, and the
robot-agnostic converter invites replication on other ROS-recorded
platforms.

\section*{Acknowledgments}
This paper is based on the results of research conducted by AIRoA (AI Robot
Association) with support from GENIAC (Generative AI Accelerator Challenge),
a project implemented by the Ministry of Economy, Trade and Industry (METI)
and the New Energy and Industrial Technology Development Organization (NEDO)
to strengthen Japan's generative AI development capabilities and accelerate
the adoption of generative AI throughout society.

\bibliographystyle{plainnat}
\bibliography{references}

\end{document}